\documentclass[11pt]{article}

\usepackage[papersize={8.5in,11in}, margin=1.0in]{geometry}
\usepackage{amsmath, amssymb, amsfonts}
\usepackage{graphicx}
\usepackage{indentfirst}
\usepackage{hyperref}
\usepackage{subcaption}
\usepackage{float}
\usepackage{setspace}
\usepackage{multicol}
\usepackage{array}
\usepackage[
    backend=biber,
    style=ieee,
    natbib=true,
    url=true, 
    doi=true,
    eprint=false,
    maxbibnames=99
]{biblatex}
\usepackage{sectsty}
\usepackage{enumitem}
\usepackage{algorithm}
\usepackage{algpseudocode}
\usepackage{listings}
\usepackage{xcolor}
\usepackage{titlesec}
\usepackage{authblk}
\usepackage{abstract}
\usepackage[title]{appendix}
\usepackage[bottom]{footmisc}
\usepackage{multirow}
\usepackage{fancyhdr}

\newcolumntype{L}[1]{>{\raggedright\let\newline\\\arraybackslash\hspace{0pt}}m{#1}}
\newcolumntype{C}[1]{>{\centering\let\newline\\\arraybackslash\hspace{0pt}}m{#1}}
\newcolumntype{R}[1]{>{\raggedleft\let\newline\\\arraybackslash\hspace{0pt}}m{#1}}

\titleformat{\section}
{\large\bfseries\centering\singlespacing}
{\arabic{section}.\space}{0pt}{}[]
\titlespacing{\section}{0pt}{0.5\baselineskip}{0pt}

\titleformat{\subsection}
{\normalsize\bfseries\singlespacing}
{\arabic{section}.\arabic{subsection}\space}{0pt}{}[]
\titlespacing{\subsection}{0pt}{0.5\baselineskip}{0pt}

\titleformat{\subsubsection}
{\normalsize\bfseries\singlespacing}
{\arabic{section}.\arabic{subsection}.\arabic{subsubsection}\space}{0pt}{}[]
\titlespacing{\subsubsection}{0pt}{0.5\baselineskip}{0pt}

\title{Scalable neural network-based blackbox optimization}
\author[1]{Pavankumar Koratikere}
\author[1,*]{Leifur Leifsson}
\affil{School of Aeronautics and Astronautics, Purdue University, West Lafayette, IN, USA}
\affil[*]{Corresponding author: leifur@purdue.edu}

\renewcommand{\abstitlestyle}[1]{}

\fancypagestyle{alim}
{\fancyhf{}\fancyhead[C]{ \textit{ This preprint is published in Structural and Multidisciplinary Optimization. Refer to the final version available at}: \url{https://doi.org/10.1007/s00158-025-04195-5}} }

\bibliography{references.bib}

\begin{document}

\date{\vspace{-5ex}}

\maketitle

\vspace{-0.2in}

\begin{abstract}
    \noindent Bayesian Optimization (BO) is a widely used approach for blackbox optimization that leverages a Gaussian process (GP) model and an acquisition function to guide future sampling. While effective in low-dimensional settings, BO faces scalability challenges in high-dimensional spaces and with large number of function evaluations due to the computational complexity of GP models. In contrast, neural networks (NNs) offer better scalability and can model complex functions, which led to the development of NN-based BO approaches. However, these methods typically rely on estimating model uncertainty in NN prediction -- a process that is often computationally intensive and complex, particularly in high dimensions. To address these limitations, a novel method, called scalable neural network-based blackbox optimization (SNBO), is proposed that does not rely on model uncertainty estimation. Specifically, SNBO adds new samples using separate criteria for exploration and exploitation, while adaptively controlling the sampling region to ensure efficient optimization. SNBO is evaluated on a range of optimization problems spanning from 10 to 102 dimensions and compared against four state-of-the-art baseline algorithms. Across the majority of test problems, SNBO attains function values better than the best-performing baseline algorithm, while requiring 40–60\% fewer function evaluations and reducing the runtime by at least an order of magnitude. An open-source implementation of SNBO is available at: \url{https://github.com/ComputationalDesignLab/snbo}
\end{abstract}

\noindent \textbf{\textit{Keywords}}: blackbox optimization, neural networks, high-dimensional optimization, Bayesian optimization, surrogate-based optimization.

\thispagestyle{alim}


\section{Introduction}\label{intro}

\noindent Many real-world optimization problems are characterized by objective functions that are expensive to evaluate and not explicitly known. Moreover, these problems are often high-dimensional in nature and evaluations can only be made at specific and a limited number of design points. Examples include aerodynamic shape optimization \citep{Liu2016}, hyperparameter optimization \citep{Feurer2019}, material design \citep{Terayama2021}, and multidisciplinary design optimization \citep{Simpson2001}. Traditional optimization techniques that rely on gradients or closed-form expressions of the objective function are not always suitable for such problems. In contrast, surrogate-based optimization (SBO) has been successfully applied to tackle these expensive problems \citep{Queipo2005}. SBO starts by building an initial surrogate model and iteratively improves it by selecting the most promising design points. These models guide the search for optimal solutions while minimizing the number of costly function evaluations, thereby significantly reducing computational expense.

Bayesian optimization (BO) \citep{Garnett2023}, also known as Efficient global optimization \citep{Jones1998}, is a widely used SBO method. BO employs a Gaussian process (GP) \citep{rasmussen2005} model due to its ability to provide both predictive estimates and associated model uncertainty measures. The process starts with a GP model built using an initial sampling points and the corresponding function observations. BO then iteratively identifies new sample points by maximizing an acquisition function, which guides the search for optimal solutions by balancing exploration and exploitation. Some of the well known acquisition functions include probability of improvement, expected improvement, and lower confidence bound \citep{Garnett2023}. This iterative refinement continues until a termination criterion is met, after which the best point within the dataset is identified as the optimal solution.

Despite its widespread application, vanilla BO is typically limited to low dimensional problems and a small number of function evaluations. This limitation arises in part because the GP model that approximates the function response as a stationary Gaussian process which does not always hold in high-dimensional settings \citep{PhanTrong2023}. For large datasets, GP's correlation matrix can become ill-conditioned, resulting in poor model uncertainty estimates that converge to the mean prediction \citep{Viana2021}. Moreover, the computational cost of training a standard GP model scales cubically with the number of samples due to the correlation matrix inversion computation \citep{snoekscalableBO}. In addition, the search space grows exponentially (due to the curse of dimensionality) which leads to large regions of uncertainty and over-exploration \citep{saasbo}.

Accordingly, various methods have been proposed to scale BO for problems with higher dimensions and larger datasets. Some methods assume that the objective function has an additive structure which can be exploited for performing BO more efficiently \citep{kandasamy2015high,gardner2017discovering,wang2018batched}. These methods often require training a large number of GP models which is not scalable for large datasets. Furthermore, the additive assumption is restrictive and may not hold for many high-dimensional problems. Another class of methods assume that the input space can be projected onto an active low-dimensional subspace where BO is conducted \citep{wang_random_embed, nayebi2019framework,letham_embedding,papenmeier2022increasing}. However, this subspace may not always exist and even if it does, identifying it efficiently is non-trivial. More recently, sparse axis-aligned subspace BO (SAASBO) \citep{saasbo} was proposed, leveraging sparsity-inducing priors in the GP model to efficiently discover low-dimensional subspaces. While promising, SAASBO does not scale well to problems requiring a large number of function evaluations. Apart from these approaches, local optimization strategies have also been explored. For instance, Eriksson et al. \citep{turbo} proposed the trust region BO (TuRBO) algorithm for solving high-dimensional problems by combining BO with trust region methods. Similarly, Wang et al. \citep{lc_mcts} utilized Monte Carlo tree search to partition the search space and then performing standard BO within each region.

Another line of research involves using a different model, such as neural networks (NNs), within BO. NNs are known to scale well for large datasets and can model complex high-dimensional functions, as suggested by the universal approximation theorem \citep{goodfellow2016deep}. A primary challenge in using NNs for BO is accurately estimating the model uncertainty so that it can easily substitute for GP. In this context, Bayesian neural networks (BNNs) \citep{Neal1996} are suitable for BO since they can explicitly provide model uncertainty estimates similar to GP models. Snoek et al. \citep{snoekscalableBO} used Bayesian linear regression within the last layer of a NN to estimate the prediction uncertainty which was further improved by Springenberg et al. \citep{springenberg}. More recently, BNNs have been studied in detail for BO, and it was found that infinite width BNNs can perform competitively for many high-dimensional problems \citep{bnn_study}. Beyond BNNs, there are other works which indirectly approximate the model uncertainty in the NN predictions. Phan-Trong et al. \citep{PhanTrong2023} employed the neural tangent kernel (NTK) \citep{ntk} to estimate confidence intervals in NN outputs. Similarly, Heiss et al. \citep{nomu} and Koratikere et al. \citep{Koratikere2023} proposed using a secondary NN to model predictive uncertainty, along with a primary NN for the function approximation.

The efficiency of these uncertainty-based BO methods is highly dependent on the accuracy of the uncertainty estimates, especially for high-dimensional problems. In practice, the uncertainty may be either underestimated or overestimated, and some of these methods can be computationally expensive as the dimension increases. However, accurate uncertainty estimation is not strictly necessary for optimizing blackbox problems. For example, Regis and Shoemaker \citep{dycors} proposed the dynamic coordinate search using response surface models (DYCORS) algorithm which refines the surrogate without relying on uncertainty estimates. DYCORS has been demonstrated on various high-dimensional problems but it uses the radial basis function as a surrogate model which suffers from a similar scalability issue as the GP model. Paria et al. \citep{paria2022greedy} introduced a greedy NN-based algorithm in which the next sample point is the current optimum of the NN model. This method emphasizes more on exploitation and does not explicitly incorporate for exploration. More recently, Borisut and Nuchitprasittichai \citep{Borisut2023} proposed a NN-based approach that uses an adaptive Latin hypercube sampling for adding new points. While this approach attempts to balance exploitation and exploration, it has only been demonstrated on low-dimensional problems.

Based on the reviewed literature, there is a lack of efficient NN-based algorithms for optimizing high-dimensional blackbox problems. To address this gap, this work introduces a scalable NN-based blackbox optimization (SNBO) method\textemdash a novel framework inspired by DYCORS and TuRBO. Rather than optimizing an acquisition function, SNBO employs a three-stage sampling strategy for adding new samples that eludes the need for uncertainty estimation. The first two stages emphasize exploration by utilizing a space-filling sequential sampling technique to generate a diverse set of points. The last stage focuses on exploitation by selecting the most promising points from this set based on the current NN predictions. Furthermore, SNBO addresses the curse of dimensionality by dynamically adjusting the search space, enabling more efficient search in high-dimensional settings. The proposed method is evaluated using a range of optimization problems and is compared against four state-of-the-art algorithms. The results demonstrate that SNBO can efficiently solve high-dimensional blackbox problems. 

The main contributions of this work are summarized as follows:
\begin{itemize}
    \item A novel and efficient NN-based approach for high-dimensional blackbox optimization that circumvents the need for model uncertainty estimation.

    \item A novel decoupled exploration–exploitation strategy that leverages sequential sampling and surrogate modeling for adding new samples.

    \item A comprehensive validation of the proposed method on a diverse set of optimization problems that elucidates the performance gains in terms of convergence rate and run time.
\end{itemize}

The remainder of this paper is organized as follows. Section \ref{sec:methods} presents the proposed algorithm in detail and provides a brief overview of the benchmark methods used for evaluation. Section \ref{sec:exp} demonstrates the effectiveness of the proposed method on a variety of analytical and real-world optimization problems, and compares it against benchmark methods. Finally, Section \ref{sec:conclusion} concludes this work, along with some recommendations for future work.


\section{Methods}\label{sec:methods}

\noindent This section provides a detailed description of the proposed scalable neural network-based blackbox optimization (SNBO) algorithm, followed by a brief overview of the four state-of-the-art methods used for benchmarking.

\subsection{Scalable neural network-based blackbox optimization}

\noindent An optimization problem involves minimizing an objective function $f\text{: }\mathbb{R}^n \to \mathbb{R}$ with respect to design variable $\mathbf{x} \in \mathbb{R}^n$, within a specified domain $\Omega$. Mathematically, it can be expressed as:
\begin{equation}
    \min_{\mathbf{x} \in \Omega} \quad f(\mathbf{x}).
    \label{eq:prob}
\end{equation}
\noindent Without the loss of generality, the domain $\Omega$ is assumed to be a unit hypercube as any bounded space can be scaled to $[0,1]^n$ space. In blackbox optimization, the internal structure and gradients of the objective function are inaccessible, and evaluations can only be performed at selected design points. Additionally, the function may be high-dimensional and computationally expensive to evaluate. The SNBO algorithm is proposed to efficiently solve these expensive high-dimensional blackbox optimization problems. It follows a similar structure as any other surrogate-based optimization (SBO) method: build a surrogate model and then adaptively refine it with new data. A NN model is employed due to its ability to scale well for large datasets and approximate complex functions, given the right architecture \citep{goodfellow2016deep}. However, the NN models do not inherently provide model uncertainty estimates that can be used to guide sampling. Estimating this model uncertainty is particularly challenging, especially for high-dimensional functions, therefore, eluding it for optimization may increase efficiency.

To this end, SNBO is designed to iteratively identify new infill points without relying on uncertainty estimates from the model. Specifically, SNBO selects new sample points in three stages. The first stage consists of creating a large candidate set by perturbing the current best point. The perturbations are sampled from a zero-mean uniform distribution whose spread dynamically adjusted based on the optimization progress to adaptively control of the search region. In the second stage, an exploration set is created by sequentially selecting points from the candidate set using a distance-based criterion. These two stages are adapted from the recently proposed fully sequential space-filling (FSSF) sampling method \citep{fssf} (cf. Section \ref{sec:exploration_set} for more details). In the final stage, the trained NN model is employed to evaluate the exploration set and select the most promising points based on the predicted function values. While the first two stages primarily focus on exploration, the final stage emphasizes exploitation. Through this three-stage strategy, SNBO effectively balances exploration and exploitation without using uncertainty estimates.

\renewcommand\arraystretch{1.25}

\begin{table}[!b]
    \centering
    \small
    \caption{Value of hyperparameters of the SNBO algorithm.}
    \begin{tabular*}{0.5\textwidth}{@{\extracolsep{\fill}}cc}
        \hline
        Parameter & Value \\
        \hline
        initial perturbation range, $r_{init}$ & 1.6 \\
        maximum perturbation range, $r_{max}$ & 1.6 \\
        minimum perturbation range, $r_{min}$ & 0.025 \\
        maximum number of success, $max_{succ}$ & 3 \\
        minimum number of success, $max_{fail}$ & $n/q$ \\
        number of exploration points, $N_{explore}$ & $nq$ \\
        perturbation probability, $p$ & $1/\sqrt{n}$ \\
        \hline
    \end{tabular*}
    \label{tab:parameter_values}
\end{table}

Algorithm \ref{alg:snbo} presents a detailed outline of the proposed SNBO method. The algorithm requires the following inputs: the blackbox objective function $f$ which can be called to evaluate sample points, the size of initial sampling plan $N_{init}$, the maximum number of function evaluations $N_{max}$, and the number of samples points to be added in each iteration $q$. The algorithm begins by initializing hyperparameters of the SNBO method (lines 2-5). The default value of these parameters are provided in Table \ref{tab:parameter_values}. Note that these values have been observed to provide good performance across a variety of problems. But tweaking these values for a specific problem might yield better results. 

The core of the SNBO algorithm is structured as a nested loop (lines 5 and 10). The outer loop is responsible for restarting the optimization process when a specific condition is met, while the inner loop handles the iterative addition of new design points. Within the outer loop, the algorithm first generates an initial sampling plan of size $N_{init}$ using the Latin hypercube sampling (LHS) \citep{Mckay2000} method, followed by evaluating the objective function $f$ at each sample point (lines 6 and 7). Subsequently, variables such as perturbation range $r$ and success/failure counters are defined. These variables are only relevant to a single optimization run, and hence, are declared within the outer loop. Note that the perturbation range $r$ governs the spread of the uniform distribution used to generate candidate set.

\begin{algorithm}[!t]
    \caption{Scalable neural network-based blackbox optimization (SNBO)}
    \label{alg:snbo}
    \begin{algorithmic}[1]
    \State{\textbf{Require}: objective function $f$, initial sample size $N_{init}$, maximum number of evaluations $N_{max}$, input dimension $n$, number of infills in each iteration $q$}
    \State{set number of evaluations $n_{evals} = 0$}
    \State{set maximum number of success and failure, $max_{succ}$ and $max_{fail}$}
    \State{set initial, maximum, and minimum value of perturbation range, $r_{init}$, $r_{max}$, and $r_{min}$}
    \State{compute number of points in exploration set, $N_{explore} = nq$}
    \While{$n_{evals} \leq N_{max}$}
        \State{$\mathcal{X} \leftarrow$ create initial sampling plan of size $N_{init}$ using LHS}
        \State{$\mathcal{Y} \leftarrow$ compute objective function $f(\mathcal{X})$}
        \State{increment $n_{evals}$ by $N_{init}$}
        \State{set $r = r_{init}, n_{succ} = 0, n_{fail} = 0$}
        \While{$n_{evals} \leq N_{max}$ \textbf{and} $r \geq r_{min}$}
            \State{$\hat{y}(\mathbf{x}) \leftarrow$ fit a NN model on $(\mathcal{X},\mathcal{Y})$ 
            dataset}
            \State{compute $\mathbf{x}_{best},y_{best}$ from $(\mathcal{X},\mathcal{Y})$ dataset}
            \State{$\mathcal{X}_{explore} \leftarrow \text{generate exploration points}(N_{explore},r,\mathbf{x}_{best},n$)  \Comment{c.f. 
            Algorithm \ref{alg:gen_explore_points}} }
            \State{$\mathcal{X}_{infill} \leftarrow$ select best $q$ points from $\mathcal{X}_{explore}$ using $\hat{y}$ \Comment{exploitation}}
            \State{$\mathcal{Y}_{infill} \leftarrow$ compute objective function $f(\mathcal{X}_{infill})$}
            \If{$min(\mathcal{Y}_{infill}) < y_{best}$}
                \State{increment $n_{succ}$ by 1 and set $n_{fail} = 0$}
            \Else
                \State{increment $n_{fail}$ by 1 and set $n_{succ} = 0$}
            \EndIf
            \If{$n_{succ}$ equal to $max_{succ}$}
                \State{set $r = $ max($2r,r_{max}$) and $n_{succ} = 0$}
            \ElsIf{$n_{fail}$ equal to $max_{fail}$}
                \State{set $r = r/2$ and $n_{fail} = 0$}
            \EndIf
            \State{append $\mathcal{X}_{infill}$ and $\mathcal{Y}_{infill}$ to $\mathcal{X}$ and $\mathcal{Y}$, respectively}
            \State{increment $n_{evals}$ by $q$}
        \EndWhile
    \EndWhile
    \State{\Return $(\mathbf{x}^*, y^*) \leftarrow$ best sample in $(\mathcal{X},\mathcal{Y})$}
    \end{algorithmic}
\end{algorithm}

Within the inner loop, the algorithm first trains the NN model using the current dataset $(\mathcal{X},\mathcal{Y})$ (line 11). Further details about the NN architecture and training procedure used in this work are provided in Section \ref{sec:neural_network}. Subsequently, the best point in the dataset is identified which is later used while adding infill points (line 12). Next, the exploration set is generated as described earlier (line 13) (cf. Section \ref{sec:exploration_set} and Algorithm \ref{alg:gen_explore_points} for the details). From this exploration set, the most promising candidate points are selected as infill points, and the objective function $f$ is evaluated at these locations (lines 14 and 15).

Following the evaluation, the counters tracking the number of consecutive successes $n_{succ}$ and failures $n_{fail}$ are updated (lines 16 to 20). An iteration is deemed successful if any of the newly evaluated infill points improve upon the current best objective value; otherwise, it is considered a failure. Based on the value of $n_{succ}$ and $n_{fail}$, the perturbation range $r$ is adjusted (lines 21 to 25). If $n_{succ}$ reaches a predefined threshold $max_{succ}$, $r$ is doubled to encourage exploration. Conversely, if $n_{fail}$ reaches the threshold $max_{fail}$, $r$ is halved to focus the search more locally. This adaptive adjustment of $r$ plays a critical role in optimization. Initially, $r$ is set such that the perturbation range overlaps with the majority of the search domain $\Omega$. As the algorithm progresses, the value of $r$ is modified to facilitate exploration and exploitation. This mechanism of adaptively changing $r$ is useful for escaping the local optima. A similar strategy is used in DYCORS \citep{dycors} and TuRBO \citep{turbo}.

The selected infill points are appended to the existing training dataset, which is then used to update the NN model in the next iteration. This iterative procedure continues until either the maximum number of function evaluations, $N_{max}$, is reached or the perturbation range $r$ falls below the minimum threshold, $r_{min}$. If $r < r_{min}$ and the evaluation budget is not exhausted, the optimization process is restarted from scratch. This restart strategy is useful to avoid local minima and is also used in various other frameworks such as TuRBO and DYCORS. Once the maximum number of function evaluations is reached, the nested loop is terminated and the best point in the dataset is returned as the final optimum.

\subsubsection{Neural networks}\label{sec:neural_network}

\noindent A NN is a machine learning model inspired by the human brain, capable of approximating nonlinear functions through interconnected layers of computational units known as neurons. According to the universal approximation theorem \citep{goodfellow2016deep}, a NN model with a sufficient number of layers and neurons can represent almost any function. Various architectures have been proposed depending on the function to be modeled. In this work, a fully-connected feedforward NN is used where each neuron in a given layer is connected to every neuron in the preceding layer. Specifically, each neuron computes a weighted combination of its inputs, followed by the application of a nonlinear activation function. Several activation functions have been introduced such as sigmoid, hyperbolic tan (tanh) and rectified linear unit (ReLU). In this work, the Gaussian error linear unit (GELU) \citep{gelu} is used since it was found to perform well for regression tasks while enabling a much smaller architecture as compared to ReLU. Moreover, GELU is smooth and differentiable across the domain which facilitates optimization during training.

For regression tasks, NNs are trained by minimizing a loss function that quantifies the difference between true and predicted values. In this work, the mean squared error is employed as the loss function. The Adaptive Moments (ADAM) optimization algorithm \citep{adam} with a learning rate of 0.001 is used to minimize the loss function. The NN weights are initialized using the He initialization method \citep{he_init}, which is recommended for activation functions like ReLU or GELU. Before training, both input and output data are standardized to zero-mean and unit variance. To avoid overfitting, NN is trained with an early stopping strategy based on the normalized root mean square error (NRMSE). The training is terminated before reaching before reaching maximum number of epochs if the NRMSE of the model on the training data is less than a tolerance. An early stopping tolerance of 0.001 is used in this work. The maximum number of epochs is set to 3,000. PyTorch \citep{pytorch}, an open-source deep learning framework, is used to build and train the network. Although, the proposed algorithm is for noise-free problems, it can be easily extended to noisy settings by 
incorporating regularization techniques during training to prevent overfitting to noise.

Since only few points are added in each iteration, the NN model is not retrained from scratch. Instead, training resumes from where it was stopped in last iteration but with the updated dataset. Additionally, the hyperparameters of the NN can impact the performance of the SNBO method, and hence, need to be determined carefully. In this work, the hyperparameters are fixed at the start and are not changed during the iteration. Some hyperparameter optimization strategies, such as BOHB \citep{falkner2018bohb}, were also explored but it did not lead to any notable improvements, primarily because the NN models employed in this work are relatively simple. Given that the maximum function evaluations $N_{max}$ is significantly larger than the initial sample size $N_{init}$, the hyperparameters are chosen such that the NN will initially overfit but the fit improves as more samples are added.

\subsubsection{Exploration point generation}\label{sec:exploration_set}

\noindent This section describes the procedure for generating the exploration set around the current best solution. The approach is inspired by the recently proposed fully sequential space-filling (FSSF) sampling method \citep{fssf}. The core idea of FSSF is simple: first create a large candidate set and sequentially select points from the set based on a criterion. Similarly, the exploration set is constructed by first generating a large number of points around current best point, and then sequentially selecting points from the set based on a distance metric. 

Algorithm \ref{alg:gen_explore_points} provides a detailed outlined of this process. The first step is to define variables such as perturbation probability $p$ and number of candidate points $N_{cand}$ (lines 2 and 3). The perturbation probability governs the number of dimensions to be perturb for each candidate point. The $N_{cand}$ is chosen to be much larger than number of exploration points $N_{explore}$ \citep{fssf}. Refer to Table \ref{tab:parameter_values} for the default values of $p$ and $N_{explore}$.

The next step involves generating the candidate set $\mathcal{X}_{cand}$ around the current best point (lines 4 to 9). For each point in $\mathcal{X}_{cand}$, the first step is to determine the number of dimensions to be perturbed, denoted by $t$, by drawing a sample from $Binom(n,p)$. The second step consists of randomly selecting $t$ dimensions that will be perturbed. In the third step, a sparse perturbation vector $\pmb{\epsilon}$ is created such that the entries corresponding to the selected dimensions are independently sampled from a uniform distribution $U[-r/2,r/2]$, while remaining entries are set to zero. Finally, $\pmb{\epsilon}$ is added to the current best point to produce the final candidate point $\mathcal{X}_{{cand}_i}$. This strategy of perturbing only a subset of dimension is shown to perform well for high-dimensional optimization \citep{Tolson2007}, and is used in DYCORS and TuRBO as well. Note that the perturbed dimensions for each candidate point is different. Moreover, the amount of perturbation along each perturbed dimension is different for each candidate point.

Once the candidate set $\mathcal{X}_{cand}$ is generated, the next step is to evaluate the distance metric for each point by computing its distance from the closest design space boundary $S$ (line 11). Finally, exploration points are generated iteratively from $\mathcal{X}_{cand}$ (lines 12 to 17). In each iteration, the candidate point corresponding to the maximum value in the distance array $\mathcal{D}$ is chosen as the next exploration point. To prevent the selection of points that are overly close to the chosen point, the distance array $\mathcal{D}$ is updated. For each point $\mathbf{x}_j$ in $\mathcal{X}_{cand}$, $\mathcal{D}_j$ is updated to $||\mathcal{X}_{{explore}_i} - \mathbf{x}_j||$ if this value is smaller than the current value of $\mathcal{D}_j$. This ensures that when $\mathcal{D}$ is maximized, points near $\mathbf{x}$ will have lower values and will not be chosen in subsequent iterations. In this way, exploration points are selected using the FSSF sampling method around the current best point.

\begin{algorithm}[t]
    \caption{generate exploration points($N_{explore},r,\mathbf{x}_{best},n$)}
    \label{alg:gen_explore_points}
    \begin{algorithmic}[1]
    \State{\textbf{Input}: number of exploration points $N_{explore}$, perturbation range $r$, best point in the dataset $\mathbf{x}_{best}$, number of dimensions $n$}
    \State{compute perturbation probability, $p = 1/\sqrt{n}$}
    \State{compute number of candidate points to be generated, $N_{cand} = 1000n + 2N_{explore}$}
    \For{$i=1,\dots,N_{cand}$}
        \State{$t \leftarrow$ determine the number of dimensions to be perturbed by drawing a sample from Binom(n,p)}
        \State{$P \leftarrow$ randomly select $t$ dimensions for perturbation}
        \State{compute perturbation vector $\pmb{\epsilon} \in \mathbb{R}^n : \epsilon_j = \begin{cases}
            \mathcal{U}[-r/2,r/2] , & \text{if $j \in P$} \\
            0, & \text{otherwise}
        \end{cases}$}
        \State{$\mathcal{X}_{{cand}_i} = \mathbf{x}_{best} + \pmb{\epsilon}$}
    \EndFor
    \State{ensure that $\mathcal{X}_{cand}$ is within unit hypercube by successive reflection}
    \State{compute distance metric array $\mathcal{D}$ where $\mathcal{D}_i = 2\sqrt{2n} \times d(\mathcal{X}_{{cand}_i},S) \text{, } \forall \text{ } i = 1,\dots,N_{cand}$}
    \For{$i = 1,\dots,N_{explore}$}
        \State{$\mathcal{X}_{{explore}_i} \leftarrow$ candidate point having maximum $\mathcal{D}$ value \Comment{exploration point}}
        \For{$j = 1,\dots,N_{cand}$}
            \State{$\mathcal{D}_j \leftarrow min(\mathcal{D}_j,||\mathcal{X}_{{explore}_i} - \mathbf{x}_j||)$ \Comment{update distance metric based on $\mathcal{X}_{{explore}_i}$}} 
        \EndFor
    \EndFor
    \State{\textbf{Output}: $\mathcal{X}_{explore}$}
    \end{algorithmic}
\end{algorithm}

\subsection{Benchmark methods}

\noindent This section briefly describes four state-of-the-art methods that are used to evaluate the proposed SNBO method. Note that all the following benchmark methods have been proposed for high-dimensional blackbox optimization.

\subsubsection{Bayesian optimization}

\noindent Bayesian optimization (BO) is a widely used technique for global optimization of blackbox functions. Similar to any SBO method, BO begins with an initial surrogate model and then adaptively refines it with new points. BO employs a Gaussian process (GP) for modeling the objective function and leverages an acquisition function, such as expected improvement (EI), to guide the selection of new points \citep{Garnett2023}. Historically, standard or vanilla BO has been considered ineffective in high-dimensional settings. However, it has been shown recently that vanilla BO can perform well for high-dimensional problems by using dimensionality-based scaling for GP lengthscale prior \citep{vanilla_bo_hvarfner}. 

Another challenge in high-dimensional BO is the optimization of an acquisition function, which often exhibits numerically vanishing values across large regions of the input domain. To address this issue, LogEI has been proposed that mitigates the numerical instability of EI and can perform well for high-dimensional cases \citep{logei}. Hence, vanilla BO is used as one of the benchmark methods in this work where the lengthscale in the GP model is scaled based on the dimension and LogEI is used as the acquisition function. The standard radial basis function (RBF) kernel is used for GP. Vanilla BO is implemented using GPyTorch \citep{gpytorch} and BoTorch \citep{botorch}.

\subsubsection{Bayesian optimization using infinite width neural networks}

\noindent Bayesian neural network (BNN) is a special type of a NN in which the model parameters are considered uncertain. Specifically, parameters are assigned some prior probability distribution which is later updated based on the data to yield posterior distribution. As a result, the output of the network is also a probability distribution that can be sampled to
compute mean predictions, along with uncertainty estimates. Recently, Li et al. \citep{bnn_study} comprehensively evaluated BNNs for BO and reported that the infinite-width BNN (IBNN) can perform well in high-dimensional settings. Theoretically, IBNN is equivalent to GP with a specific kernel function that depends on the hyperparameters of the network and can model non-stationary GPs \citep{ibnn}. In this work, IBNN coupled with LogEI is used as one of the benchmarks. All the hyperparameter settings follow those provided in the original study. IBNN is implemented using BoTorch and the original source code available at: \url{https://github.com/yucenli/bnn-bo}.

\subsubsection{Trust region Bayesian optimization}

\noindent Trust region BO (TuRBO) is a state-of-the-art method for high-dimensional BO that combines classical trust-region (TR) methods with BO \citep{turbo}. Specifically, TuRBO considers a hyper-rectangle TR centered around current best point and looks for new infill points within this region only. The size of the TR is computed based on the lengthscale of the GP model along each dimension. Within TR, TuRBO generates a perturbation-based candidate set using a Sobol' sequences and then uses Thompson sampling \citep{Garnett2023} to determine next set of infill points. Moreover, TR can grow or shrink depending on the optimization progress. In this way, TuRBO employs local optimization to tackle high-dimensional problems. Table \ref{tab:comparison} provides a detailed comparison between TuRBO and SNBO, refer to Section \ref{sec:comparison} for more details. Eriksson et al. also proposed a multiple TR version of the method but in this work only one TR is used. All implementation settings follow those in the original study. TuRBO is implemented using GPyTorch and the original source code available at: \url{https://github.com/uber-research/TuRBO}.

\subsubsection{Dynamic coordinate search using response surface models}

\noindent Regis and Shoemaker \citep{dycors} introduced dynamic coordinate search using response surface model (DYCORS) for solving high-dimensional blackbox optimization problems. DYCORS integrates local metric stochastic response surface model \citep{lmsrs} with dynamically dimensioned search algorithm \citep{Tolson2007}. It starts with constructing a RBF model and then iteratively refine it with new points. In each iteration, perturbation-based candidate set is created using zero-mean normal distribution around the current best point. The variance of the distribution can increase or decrease depending on the optimization progress. Next, a weighted score function is used to select a infill point from the set. The score function combines a distance metric and surrogate model prediction to balance exploration and exploitation. DYCORS employs local optimization strategies to solve high-dimensional problems. DYCORS and SNBO are compared in Table \ref{tab:comparison}, refer to Section \ref{sec:comparison} for more details. All the algorithmic settings are consistent with those presented in the original paper. This work uses an open-source implementation of DYCORS available at: \url{https://github.com/aquirosr/DyCors}.

\subsubsection{Comparison of SNBO against the benchmark methods}\label{sec:comparison}

\renewcommand\arraystretch{1.5}

\begin{table}[!b]
    \small
    \centering
    \caption{A comparison between SNBO and other state-of-the-art benchmark methods.}
    \begin{tabular*}{\textwidth}{@{\extracolsep{\fill}}L{2cm}C{2cm}C{2cm}C{2.5cm}C{2.5cm}C{2.5cm}}
        \hline
        - & BO+LogEI & IBNN & TuRBO & DYCORS & SNBO \\
        \hline
        Model & GP & IBNN & GP & RBF & NN \\
        Uncertainty & Yes & Yes & Yes & No & No \\
        Search region & Global & Global & Local & Local & Local \\ 
        Acquisition method & Optimization & Optimization & Sampling & Sampling & Sampling \\
        Acquisition function & LogEI & LogEI & Thompson sampling & Weighted score & Sequential sampling \& surrogate model \\
        Perturbation method & - & - & Sobol' sequence & Normal dist. & Uniform dist. \\
        Number of dimensions to perturb & - & - & Constant & Decreases with iteration & Constant \\
        Restart capability & No & No & Yes & Yes, with previous best & Yes \\
        \hline
    \end{tabular*}
    \label{tab:comparison}
\end{table}

\noindent  Table \ref{tab:comparison} summarizes key differences between the proposed SNBO method and the benchmark methods. The search region for TuRBO, DYCORS, and SNBO is centered around the current best point while BO+LogEI and IBNN search within the entire domain. Also, BO+LogEI and IBNN optimize an acquisition function to find the next set of infill points. The other three methods rely on creating a set of points and then selecting the best performing point as the next infill. It should be noted that optimizing an acquisition function is numerically challenging, especially for high-dimensional problems. Moreover, there is no provision for restarting the optimization in BO+LogEI or IBNN. The capability to restart is required in TuRBO, DYCORS, and SNBO to avoid getting trapped in a local optimum. Note that DYCORS restarts with an initial sampling plan that contains the best point found in previous run but TuRBO and SNBO restart without any information from previous runs.

While SNBO shares certain similarities with TuRBO and DYCORS, there are several key differences that distinguish it from these methods. Both TuRBO and DYCORS rely on a single criterion for finding next infill point, but SNBO uses two separate criteria. This decoupling of exploration and exploitation allows greater flexibility in choice of criteria, enabling each to be tailored independently to better suit the objective function. The perturbation added in DYCORS is based on the zero-mean normal distribution which tends to concentrate candidate points near the current best solution. TuRBO generates perturbation using a Sobol' sequence, but does not follow any distribution as such. Meanwhile, SNBO uses zero-mean uniform distribution which results in more diverse perturbations and not too close to the best point.

Finally, all three methods use the $Binom(n,p)$ distribution to determine the number of dimensions to be perturbed but the probability of success ($p$) is different across the methods. In TuRBO and DYCORS, the initial value of $p$ is $min(1,20/n)$, but in DYCORS $p$ monotonically decreases towards zero with each iteration while in TuRBO it is held constant. In SNBO, $p$ is set to $1/\sqrt{n}$ (Table \ref{tab:parameter_values}) and is constant throughout the iteration. As a result, for TuRBO and DYCORS, the average number of dimensions to be perturbed will be 20 for $n>20$, while it will be $\sqrt{n}$ for SNBO. Moreover, for DYCORS, decreasing the value of $p$ with each iteration, reduces the scope of exploration. In summary, SNBO has more exploration capability in comparison to other two methods, especially in high-dimensional settings.


\section{Numerical Experiments}\label{sec:exp}

\noindent This section outlines various analytical and real world optimization problems that are used to demonstrate the SNBO method, and compare its results against other benchmarks. All methods are run ten times to account for randomness and for each run, the initial sampling plan is same for all methods. For all test problems, the initial sampling plan consists of $2\times n$ number of samples and the number of infills added in each iteration ($q$) is set to 1. All experiments are conducted on 32 CPU cores and an NVIDIA A100 GPU. Note that DYCORS is a CPU-only method and does not utilize GPU acceleration. The convergence history over ten runs is visualized as a convergence band. The center-line in the band represents median, while the upper and lower limit, denote the $75^{th}$ and $25^{th}$ percentiles, respectively. Additionally, for each test problem, the best, median, and worst result obtained by each method is tabulated. The median total time to complete a single optimization run is also reported.

\subsection{Analytical problems}

\begin{figure}[!b]
    \centering
    \includegraphics[width=\textwidth]{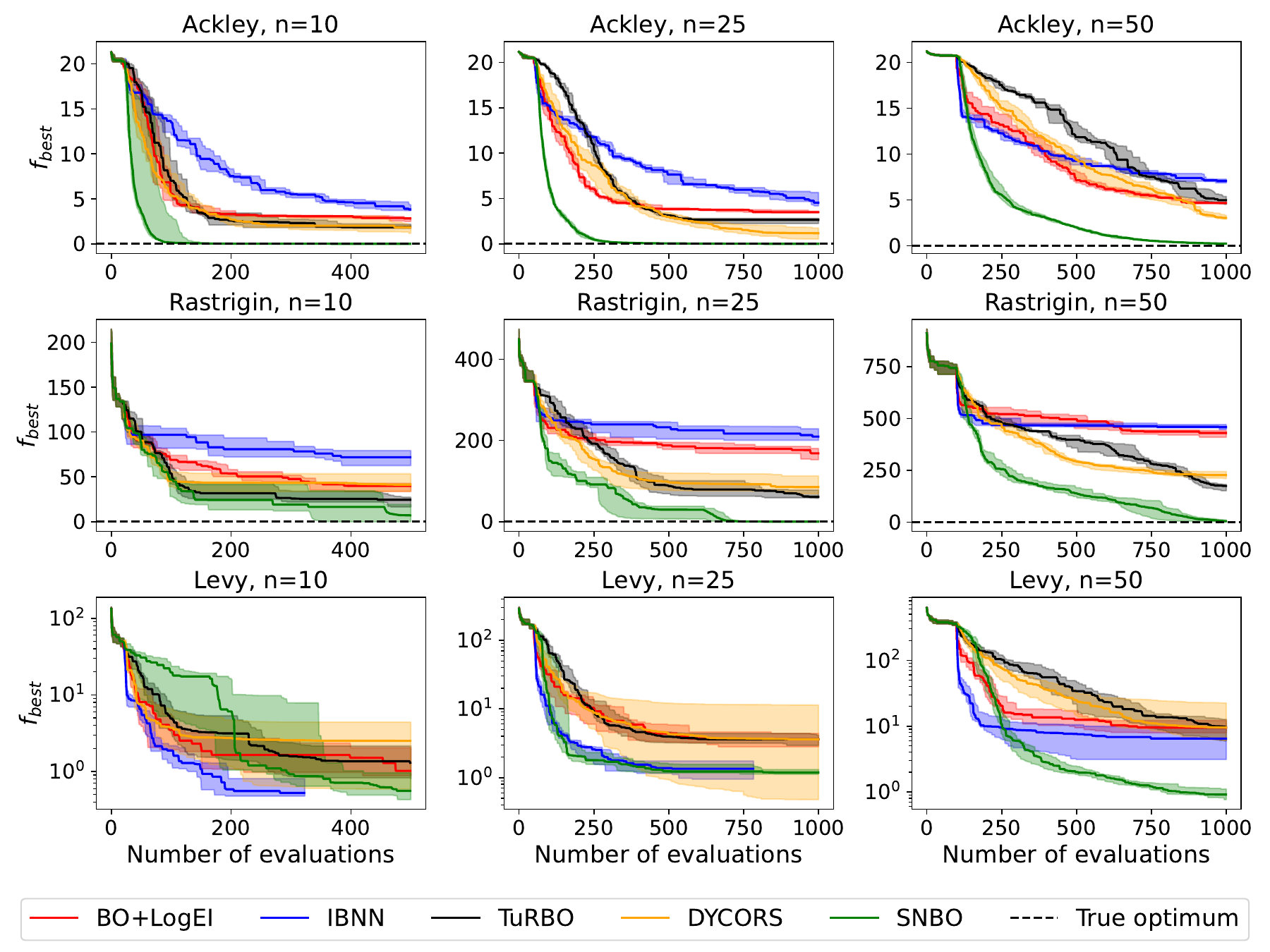}
    \caption{Convergence for the Ackley, Rastrigin, and Levy function optimization.}
    \label{fig:ackley_rastrigin_levy}
\end{figure}

\noindent This section demonstrates the SNBO method using a set of well-established analytical problems, and compares its performance against the benchmark methods. Specifically, the Ackley, Rastrigin, and Levy functions are used, each evaluated at three different dimensionalities: 10, 25, and 50. This yields a total of nine test problems. The exact function details and corresponding design variable bounds can be found at: \url{https://www.sfu.ca/~ssurjano/optimization.html}.

The initial sampling plan consists of 20, 25, and 100 samples for the 10, 25, and 50-dimensional cases, respectively. For the 10-dimensional problems, the maximum number of function evaluations is set to 500 and the NN model used in SNBO consists of two hidden layers, each containing 128 neurons. For the higher dimensional problems, the maximum function evaluations is increased to 1000, and the NN architectures consist of two hidden layers with 256 neurons each.

Figure \ref{fig:ackley_rastrigin_levy} presents the optimization histories for all methods across the nine problems. Overall, SNBO consistently outperforms the competing methods across all functions and dimensional settings. It demonstrates a strong sample efficiency by reaching near-optimal solutions within the allowed evaluation budget, while other methods exhibit slower convergence and often plateau prematurely.

In particular, for the 10-dimensional Rastrigin and Levy functions, methods such as TuRBO and IBNN initially show comparable performance to SNBO. However, as the iterations progress, these methods struggle to make further improvements and tend to stagnate near suboptimal values. This highlights SNBO’s ability to continue exploring effectively and refine solutions even in later stages of the optimization process.

\renewcommand\arraystretch{1.2}

\begin{table}[!b]
    \small
    \centering
    \caption{Summary of the results of the Ackley function optimization.}
    \small
    \begin{tabular*}{0.85\textwidth}{@{\extracolsep{\fill}}ccccccc}
         \multicolumn{2}{c}{Problem} & BO+LogEI & IBNN & TuRBO & DYCORS & SNBO \\ 
         \hline 
         \multirow{4}*{Ackley 10D} 
          & Best        & 1.8795 & 3.1837 & 1.4589 & 0.1277 & \textbf{0.0003} \\
          & Median      & 2.8269 & 3.8274 & 1.8637 & 1.8384 & \textbf{0.0007} \\
          & Worst       & 3.3052 & 4.7828 & 2.4854 & 3.0440 & \textbf{0.0016} \\
          & Time (min)  & 8.6525 & 14.5388 & 2.8148 & \textbf{0.9704} & 4.7337 \\
         \hline 
         \multirow{4}*{Ackley 25D} 
          & Best        & 2.6486 & 3.5646 & 2.0325 & 0.1178 & \textbf{0.0026} \\
          & Median      & 3.4927 & 4.5339 & 2.6522 & 1.1728 & \textbf{0.0044} \\
          & Worst       & 3.7591 & 6.2643 & 3.2372 & 2.8151 & \textbf{0.0067} \\
          & Time (min)  & 23.5414 & 30.7608 & 6.1915 & 10.7271 & \textbf{1.8690} \\
         \hline
         \multirow{4}*{Ackley 50D} 
          & Best        & 4.4201 & 6.1432 & 4.1028 & 2.3671 & \textbf{0.1598} \\
          & Median      & 4.6586 & 7.0465 & 4.9635 & 3.0289 & \textbf{0.2341} \\
          & Worst       & 4.8899 & 8.3757 & 6.0896 & 5.2317 & \textbf{0.2986} \\
          & Time (min)  & 35.8961 & 29.9924 & 5.9809 & 32.3933 & \textbf{1.8313} \\
         \hline
    \end{tabular*}
    \label{tab:ackley_summary}
\end{table}

\begin{table}[!t]
    \small
    \centering
    \caption{Summary of the results of the Rastrigin function optimization.}
    \small
    \begin{tabular*}{0.85\textwidth}{@{\extracolsep{\fill}}ccccccc}
         \multicolumn{2}{c}{Problem} & BO+LogEI & IBNN & TuRBO & DYCORS & SNBO \\ 
         \hline 
         \multirow{4}*{Rastrigin 10D} 
          & Best        & 28.5735 & 56.8543 & 12.5938 & 21.8891 & \textbf{0.0000} \\
          & Median      & 39.9104 & 71.9459 & 24.4171 & 41.3220 & \textbf{7.0659} \\
          & Worst       & 48.0620 & 81.0138 & 40.1375 & 68.6557 & \textbf{30.2466} \\
          & Time (min)  & 15.7705 & 14.9337 & 2.2846 & \textbf{0.7605} & 1.4086 \\
         \hline 
         \multirow{4}*{Rastrigin 25D} 
          & Best        & 141.5428 & 186.3649 & 53.3250 & 49.9506 & \textbf{0.0008} \\
          & Median      & 168.2384 & 209.5010 & 61.3407 & 85.0915 & \textbf{0.0161} \\
          & Worst       & 186.9712 & 249.2235 & 97.5356 & 125.8409 & \textbf{16.6427} \\
          & Time (min)  & 87.2152 & 23.4040 & 5.2405 & 10.6378 & \textbf{1.8262} \\
         \hline
         \multirow{4}*{Rastrigin 50D} 
          & Best        & 372.8138 & 422.0618 & 115.7125 & 149.6187 & \textbf{1.1421} \\
          & Median      & 429.8198 & 457.9174 & 174.5929 & 226.6926 & \textbf{4.5042} \\
          & Worst       & 462.3454 & 483.4559 & 223.9614 & 287.3534 & \textbf{138.3703} \\
          & Time (min)  & 92.7202 & 16.8874 & 5.8285 & 32.1038 & \textbf{2.4456} \\
         \hline
    \end{tabular*}
    \label{tab:rastrigin_summary}
\end{table}

Tables \ref{tab:ackley_summary}, \ref{tab:rastrigin_summary}, and \ref{tab:levy_summary} summarize the final optimization results obtained by each method across all test problems. As observed earlier, SNBO consistently outperforms the benchmark methods across most problems, achieving values close to the global optimum while also maintaining the lowest median runtime. For the 10-dimensional problems, the median runtime of methods such as DYCORS is comparable to that of SNBO. However, as the dimensionality increases, the computational cost of other methods rise significantly, whereas SNBO maintains competitive runtimes which demonstrates its strong scalability.

These results highlight the computational efficiency of SNBO, particularly in high-dimensional settings. Notably, SNBO achieves this performance without explicitly modeling predictive uncertainty for NN. Another contributing factor to SNBO's lower runtime is its use of a sampling-based acquisition strategy rather than direct optimization of an acquisition function, which can be computationally demanding and less stable in high-dimensional spaces. Finally, the results validate the effectiveness of the decoupled exploration and exploitation strategy that can enable a more customized search process.

\renewcommand\arraystretch{1.2}

\begin{table}[t]
    \small
    \centering
    \caption{Summary of the results of the Levy function optimization.}
    \small
    \begin{tabular*}{0.85\textwidth}{@{\extracolsep{\fill}}ccccccc}
         \multicolumn{2}{c}{Problem} & BO+LogEI & IBNN & TuRBO & DYCORS & SNBO \\ 
         \hline 
         \multirow{4}*{Levy 10D} 
          & Best        & 0.5226 & 0.3625 & 0.1044 & \textbf{0.0000} & 0.2575 \\
          & Median      & 1.0190 & \textbf{0.5242} & 1.3005 & 2.5122 & 0.5587 \\
          & Worst       & 4.3810 & \textbf{1.3226} & 3.1643 & 10.4318 & 13.9091 \\
          & Time (min)  & 8.2796 & 5.8040 & 3.0069 & 0.9697 & \textbf{0.7498} \\
         \hline 
         \multirow{4}*{Levy 25D} 
          & Best        & 1.8162 & 0.6760 & 1.3845 & \textbf{0.0015} & 0.2760 \\
          & Median      & 3.6132 & 1.3451 & 3.6134 & 3.6027 & \textbf{1.1873} \\
          & Worst       & 6.8601 & 2.8132 & 11.5409 & 20.7904 & \textbf{1.6293} \\
          & Time (min)  & 24.5297 & 6.8657 & 6.5551 & 10.7491 & \textbf{0.7281} \\
         \hline
         \multirow{4}*{Levy 50D} 
          & Best        & 5.1076 & 2.0341 & 3.2364 & 2.7265 & \textbf{0.6402} \\
          & Median      & 9.4865 & 6.4682 & 9.6320 & 9.5249 & \textbf{0.9122} \\
          & Worst       & 13.7498 & 11.0162 & 32.3371 & 37.8121 & \textbf{1.7095} \\
          & Time (min)  & 23.2451 & 12.7638 & 7.1248 & 32.1158 & \textbf{1.0697} \\
         \hline
    \end{tabular*}
    \label{tab:levy_summary}
\end{table}

\subsection{Real-world optimization problems}

\noindent This section evaluates SNBO on two real-world optimization problems and compares it against other benchmark methods. Specifically, rover trajectory optimization and Half-Cheetah optimization problems are used. The rover problem, originally proposed by Wang et al. \citep{wang2018batched}, has become a widely used benchmark for assessing the scalability and efficiency of high-dimensional optimization algorithms. The objective of the problem is to determine an optimal path for a rover to move from a specified start point $\mathbf{x}_s$ to a goal point $\mathbf{x}_g$, within a unit square domain, while minimizing the total travel cost.

\begin{figure}[!t]
    \centering
    \includegraphics[width=0.5\textwidth]{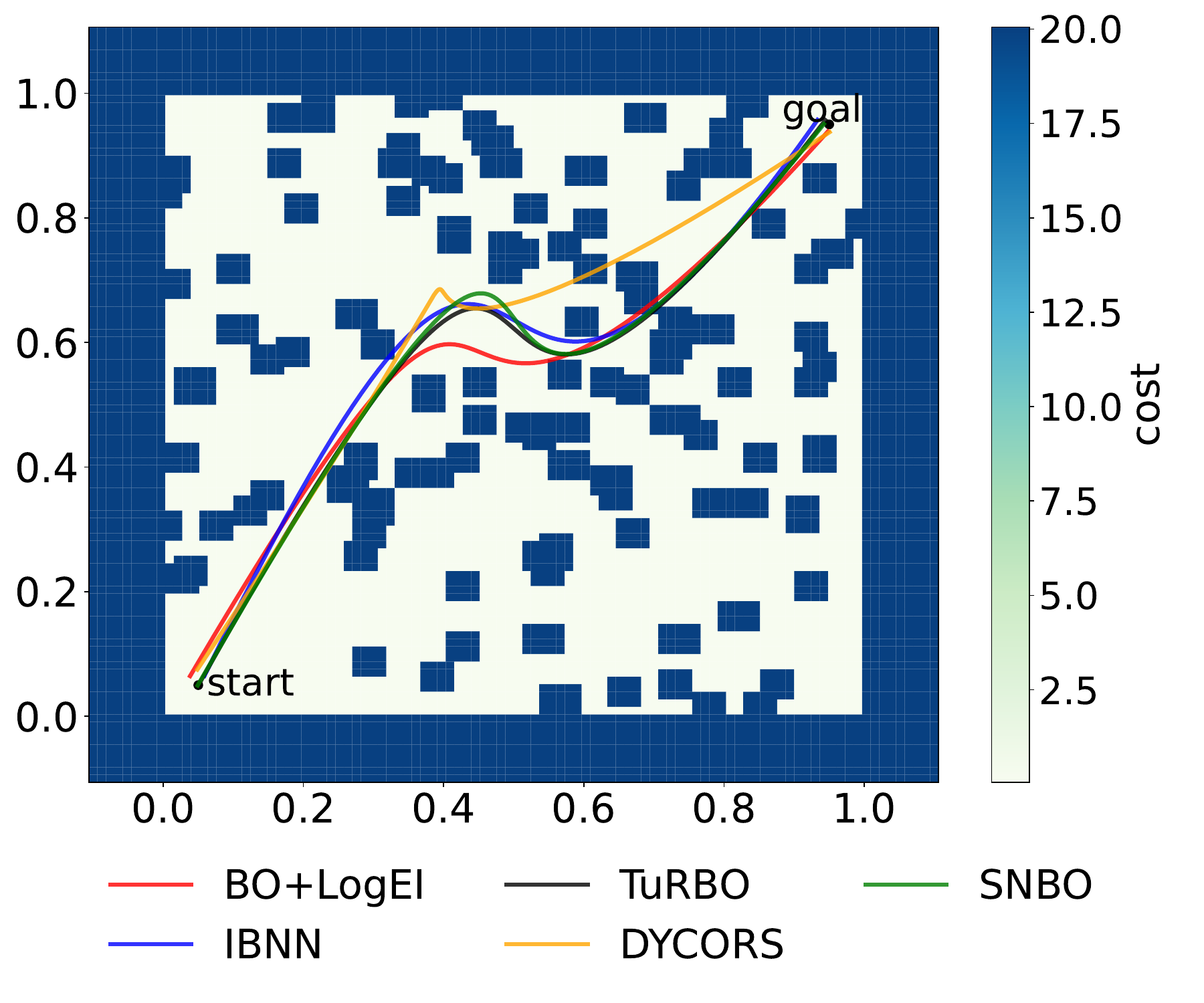}
    \caption{Best trajectory found by each method for maneuvering the rover from start to end.}
    \label{fig:rover_path}
\end{figure}

Figure \ref{fig:rover_path} illustrates a cost contour for maneuvering the rover in the domain. The lighter regions correspond to lower cost areas, while darker regions are high cost zones. The rover's path is parameterized using a B-spline curve that is defined using 50 control points. As a result, there are 100 design variables in this problem since each control point is defined by the x- and y-coordinates. The optimizer has to tweak these control points so that the path originates from the $\mathbf{x}_s$ point, maneuvers through low cost regions, and ends at the $\mathbf{x}_g$ point. Mathematically, the objective function $f(\mathbf{x})$ is formulated as $c(\mathbf{x}) + \lambda \Big[ || \mathbf{x}_{1,2} - \mathbf{x}_s ||_1 + || \mathbf{x}_{99,100} - \mathbf{x}_g ||_1 \Big] + b$, where $c$ is the total trajectory cost and $b$ is a constant offset. The $\mathbf{x}_{1,2}$ denotes the first and second design variable, while $\mathbf{x}_{99,100}$ refers to last two variables. The second term penalizes for not starting and ending at the predefined location, while $\lambda$ acts like a penalty multiplier. As described in the original formulation, the values of $\lambda$ and $b$ are set to 10 and 5, respectively. The total trajectory cost $c(\mathbf{x})$ is computed by integrating the cost of individual steps along the trajectory, refer to Wang et al. \citep{wang2018batched} for more details.

The Half-Cheetah problem is a widely used reinforcement learning (RL) benchmark that has also been used for evaluating high-dimensional optimization methods. The problem consists of a two-dimensional cheetah-like robot that needs to be trained to run forward. The robot consists of nine body parts that are connected through eight joints, out of which only six can be actuated. The state space consists of 17 variables, while the action space consists of 6 variables corresponding to the actuated joints. 

The objective of this problem is to learn a policy that selects suitable actions given the current state, with the aim of maximizing the cumulative reward over time. Typically, a linear policy is assumed for the sake of simplicity which results in 102 dimensions.  The total reward associated with a given policy is computed by simulating the robot within its environment for a predefined number of time steps. The robot and its environment, are implemented using gymnasium \citep{gym}, an open-source RL library. For a detailed description for reward function computation, refer to the gymnasium documentation\footnote{\url{https://gymnasium.farama.org/environments/mujoco/half_cheetah/}}.

The initial sampling plan consists of 200 and 204 points for the rover and half cheetah problem, respectively. For both problems, the NN model consists of two hidden layers with 256 neurons in each layer and the maximum number of evaluations is set to 2000. Note that BO+LogEI is terminated after 1500 function evaluations due to the high computational cost for training the GP model and optimizing the acquisition function.

\begin{figure}[!t]
    \centering
    \includegraphics[width=0.7\textwidth]{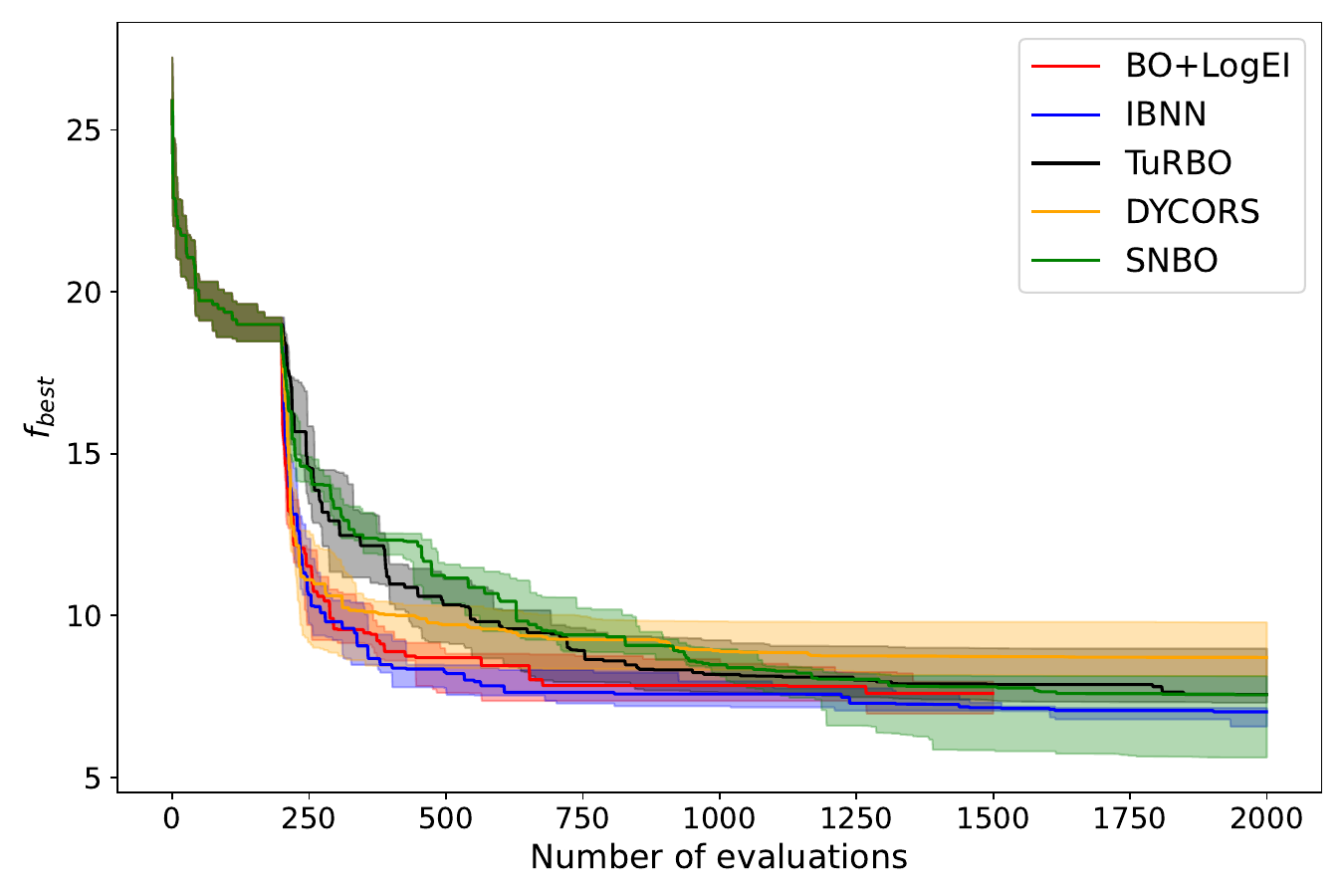}
    \caption{Convergence history for the rover trajectory optimization problem.}
    \label{fig:rover_conv}
\end{figure}

Figure~\ref{fig:rover_conv} shows convergence history for the rover trajectory optimization problem. BO+LogEI, IBNN, and DYCORS exhibit fast initial convergence but show limited improvement in later stages. TuRBO and SNBO have similar convergence rate throughout the optimization but SNBO does provide better optimum values for some of the runs.

A detailed summary of the rover optimization problem is provided in Table~\ref{tab:real}. SNBO achieves the best final objective value of 5.36 and it also reports the second-best median value — slightly behind IBNN. Figure~\ref{fig:rover_path} visualizes the best optimum path found by each method. Both SNBO and TuRBO obtain similar trajectories that begin at the designated start location, traverses primarily through low-cost regions, and terminate at the goal location. While IBNN also generates a similar trajectory but it is marginally longer which results in slightly higher final cost.

\renewcommand\arraystretch{1.25}

\begin{table}[t]
    \small
    \centering
    \caption{Summary of the results of the real-world optimization problems.}
    \small
    \begin{tabular*}{\textwidth}{@{\extracolsep{\fill}}ccccccc}
         \multicolumn{2}{c}{Problem} & BO+LogEI & IBNN & TuRBO & DYCORS & SNBO \\ 
         \hline
         \multirow{4}*{Rover problem 100D} 
          & Best        & 6.11 & 6.187 & 5.39 & 6.59 & \textbf{5.36} \\
          & Median      & 7.60 & \textbf{7.03} & 7.56 & 8.71 & 7.55 \\
          & Worst       & 8.37 & \textbf{7.63} & 9.67 & 12.87 & 10.36 \\
          & Time (min)  & 121.86 & 30.01 & 13.12 & 161.12 & \textbf{6.95} \\
         \hline
         \multirow{4}*{Half Cheetah Problem 102D}
         & Best  & -1635.17 & -2462.87 & -1184.40 & 160.71 & \textbf{-3493.66} \\
         & Median & -578.17 & \textbf{-1211.98} & -424.28 & 657.39 & -832.52 \\
         & Worst  & -47.26 & \textbf{-858.45} & 222.79 & 1495.42 & 131.45 \\
         & Time (min) & 92.22 & 51.29 & 21.37 & 119.93 & \textbf{13.81} \\
    \hline
    \end{tabular*}
    \label{tab:real}
\end{table}

\begin{figure}[!t]
    \centering
    \includegraphics[width=0.7\textwidth]{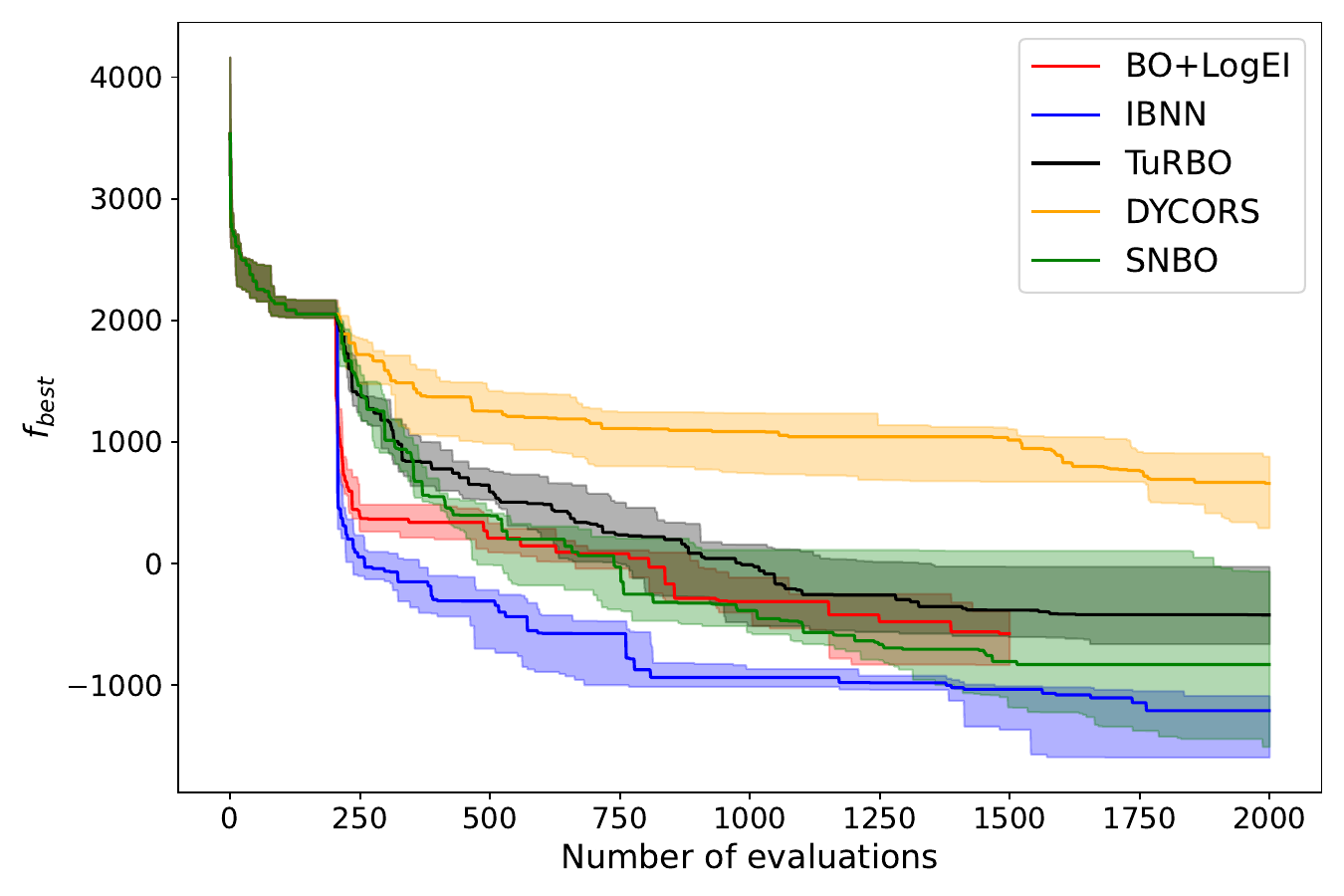}
    \caption{Convergence history for the Half cheetah problem}
    \label{fig:halfcheetah_conv}
\end{figure}

Figure \ref{fig:halfcheetah_conv} illustrates the convergence behavior of all methods for the Half-Cheetah problem. SNBO performs better than other methods, except for IBNN which has a faster convergence rate. However, SNBO still obtains the best optimum value as outlined in Table \ref{tab:real}, while achieving second best median value and atleast an order of magnitude faster runtime.

For these two real-world optimization problems, SNBO does not outperform other methods but has a comparable performance. This can be attributed to certain limitations of the proposed approach. Like other local optimization strategies, such as DYCORS and TuRBO, SNBO is highly sensitive to its hyperparameter settings. In particular, convergence behavior is strongly influenced by the evolution of the perturbation range $r$. If the value of $r$ is not appropriate during any stage of the optimization, the convergence can significantly deteriorate. Moreover, other hyperparameters such as the number of exploration points $N_{explore}$ and the perturbation probability $p$ also affect the performance. Furthermore, given the large number of evaluations, optimal set of hyperparameters for the NN model might change as optimization progresses. Thus, incorporating adaptive hyperparameter tuning heuristics may be necessary to maintain efficiency throughout the optimization process. Addressing even some of these limitations can further improve SNBO’s performance in complex, high-dimensional settings.


\section{Conclusion}\label{sec:conclusion}

\noindent This work introduces a novel NN-based approach for efficiently solving blackbox optimization problems. The proposed method, referred to as scalable NN-based blackbox optimization (SNBO), constructs an initial NN model and iteratively refines it with new sample points. As the model uncertainty estimation for NN predictions is non-trivial--particularly in high-dimensional spaces--SNBO employs an innovative sampling strategy that avoids reliance on such estimates.

Specifically, SNBO uses a three-stage sampling procedure designed to balance exploration and exploitation. The first two stages emphasize exploration by generating candidate points in the vicinity of the current best solution through a sequential sampling approach. The third stage focuses on exploitation by selecting the most promising points from the generated set based on the NN predictions. Importantly, this sampling strategy avoids the use of acquisition function, which are often numerically challenging to optimize in high-dimensional settings. Furthermore, SNBO dynamically adjusts the sampling region throughout the optimization process to alleviate the challenges associated with  high-dimensional search spaces.

The proposed algorithm is evaluated on 10, 25, and 50-dimensional analytical problems, as well as on two real-world optimization tasks. Its performance is compared against four contemporary high-dimensional methods. The experimental results demonstrate strong sample efficiency and scalability for SNBO across most problems, as well as a low computational time. Future work will explore extending SNBO to handle constrained optimization problems. Additionally, similar to TuRBO, SNBO can be modified to support multiple concurrent local optimizations which can further enhance sampling efficiency.

\section*{Acknowledgments}

\noindent This work was supported in part by the U.S. National Science Foundation (NSF) award number 2223732 and by the Icelandic Centre for Research (RANNIS) award number 239858.


\printbibliography

\end{document}